\documentclass{article}

\usepackage[final,nonatbib]{neurips_2019}

\usepackage[utf8]{inputenc} % allow utf-8 input
\usepackage[T1]{fontenc}    % use 8-bit T1 fonts
\usepackage{hyperref}       % hyperlinks
\usepackage{url}            % simple URL typesetting
\usepackage{booktabs}       % professional-quality tables
\usepackage{amsfonts}       % blackboard math symbols
\usepackage{nicefrac}       % compact symbols for 1/2, etc.
\usepackage{microtype}      % microtypography
\usepackage{amsmath}
\usepackage{multirow}
\usepackage{subcaption}

\usepackage{mathtools,graphicx}

\makeatletter
\newcommand{\parop}{\DOTSB\parop@\slimits@}
\newcommand{\parop@}{\mathop{\vphantom{\sum}\mathpalette\delimiter@to@op\Vert}}
\newcommand{\delimiter@to@op}[2]{%
  \begingroup
  \sbox\z@{$#1\sum$}%
  \vcenter{\hbox{\resizebox{!}{\ht\z@}{$\m@th#2$}}}%
  \endgroup
}
\makeatother

\title{Representation Learning of EHR Data via Graph-Based Medical Entity Embedding}

\author{
    Tong Wu$^{1,2}$, Yunlong Wang$^{1\ast}$, Yue Wang$^{1}$, Emily Zhao$^{1}$, Yilian Yuan$^{1}$, Zhi Yang$^{2}$ \\
    $^{1}$ Advanced Analytics, IQVIA Inc., Plymouth Meeting, PA 19462 \\
    $^{2}$ Biomedical Engineering, University of Minnesota Twin Cities, Minneapolis, MN 55455
}

\begin{document}

\maketitle

\begin{abstract}
Automatic representation learning of key entities in electronic health record (EHR) data is a critical step for healthcare informatics that turns heterogeneous medical records into structured and actionable information.
Here we propose \texttt{ME2Vec}, an algorithmic framework for learning low-dimensional vectors of the most common entities in EHR: medical services, doctors, and patients.
\texttt{ME2Vec} leverages diverse graph embedding techniques to cater for the unique characteristic of each medical entity.
Using real-world clinical data, we demonstrate the efficacy of \texttt{ME2Vec} over competitive baselines on disease diagnosis prediction.
\end{abstract}

\section{Introduction}
\vspace{-5pt}

Recent years have seen an explosion in the growth of electronic health record (EHR) data.
One major challenge of \emph{representation learning} in EHR comes from the heterogeneity of the various medical entities that compose EHR data, including diagnoses, prescriptions, medical procedures, doctor profiles, and patient demographics, etc.
Furthermore, the relational and longitudinal structure of organizing medical entities in patient medical records (or \emph{patient journeys}) makes it more challenging to design effective and scalable representation learning algorithms:
a patient may visit one or more clinical sites multiple times with irregular time intervals, with each visit generating a varying number of medical services (diagnoses, prescriptions, or procedures) from possibly different doctors.

To address the above challenges, Choi et al. leveraged the multilevel structure of EHR data where diagnosis codes categorize treatment codes within each visit and learned a multilevel medical embedding for predictive healthcare \cite{choi2018mime,choi2019graph}.
Though being effective, their approaches do not consider the temporal characteristics unique to individual medical services, hence cannot properly address the irregular time intervals of visits that are pervasive in patient journeys.
Some recent works treated medical services in patient journeys as words in documents \cite{zhu2016measuring,cai2018medical}, and since similar words (medical services) tend to share similar contexts, word embedding techniques such as \texttt{Word2Vec} \cite{mikolov2013distributed} can be adopted to train the embedding vectors of medical services.
In this approach, a key design choice is the length of context window, or temporal scope, which should preferably vary for different medical services.
As manually specifying the temporal scope for each service is infeasible, an attention mechanism is proposed in \cite{cai2018medical} to derive a ``soft'' temporal scope for each service, where the attention coefficients can be trained jointly with the parameters in \texttt{Word2Vec}.
A caveat of this approach is that the context window has to be sufficiently large for medical services with long time spans of influence, which would significantly elevate the computational overheads for all services.
Some recent works explored the study of patient similarity, which is believed to be an enabling technique for various healthcare applications such as cohort analysis and personalized medicine \cite{zhu2016measuring,sharafoddini2017patient,suo2018multi,huai2018uncorrelated}.

In this work, we propose a graph-based, hierarchical medical entity embedding framework \texttt{ME2Vec} that can address the aforementioned challenges.
At the service level, we propose to characterize the importance of heterogeneous medical services with their co-occurrence frequencies.
Namely, important services are typically infrequent in patient journeys, hence their co-occurrence frequencies with other services are smaller than those of routine services.
With a proximity-preserving embedding approach, important services with small co-occurrence frequencies will be far away from other services in the embedding space, thus emphasizing their importance via ``spatial isolation''.
At the doctor and patient level, a fundamental principle we adhere to is ``\emph{It's what you do that defines you}'', which empowers the interpretability of embeddings.
For example, the embedding vector of a doctor is solely calculated from the doctor's conducted medical services.
To preserve the network proximities of patient vertices w.r.t. both doctor and service vertices, we develop a method called \emph{duplication \& annotation} that can convert an attributed multigraph to a simple graph without loss of structural information, to which efficient and scalable graph embedding techniques can be applied with ease.

Overall, \texttt{ME2Vec} provides a unified solution of embedding medical entities, thus can serve as a general-purpose representation learning algorithm for EHR data.

\section{Methods}
\vspace{-5pt}

\noindent\textbf{Service Embedding}$\quad$
We create the graph of medical services $\mathcal{G}_{svc}=(\mathcal{S}, \mathcal{E}_{svc})$, where $\mathcal{S}=\{s_1, s_2, \dots, s_{|\mathcal{S}|}\}$ is the set of medical services, and $\mathcal{E}_{svc}$ is the set of edges connecting medical services.
The weight of $e_{ij}$ denotes the co-occurrence frequency of services $s_i$ and $s_j$.
To obtain the adjacency matrix $\mathbf{A}_{svc}  \in \mathbb{R}^{|\mathcal{S}| \times |\mathcal{S}|}$, we use a $T$-day context window to traverse all patient journeys with no overlap.
At each location, we update $\mathbf{A}_{svc}$ with the count of the occurrence of each unique pair of medical services appeared within the $T$ days of the current window.
Note that the co-occurrence frequencies of services are summed over different patients, thus reflecting a generalized knowledge of the time intervals between medical services, which can enhance the robustness and transferability of the learned service embedding.

To embed medical services, we first obtain the adjacency matrix $\mathbf{A}_{svc}$ from patient journeys and use it to generate biased random walks, then optimize the embeddings of medical services by maximizing the probability of each service ``seeing'' its neighbors in the walks via \texttt{Word2Vec}.

\noindent\textbf{Doctor Embedding}$\quad$
We note that medical services conducted by a doctor exhibit patterns that are consistent with the doctor's primary specialty.
For example, prescriptions and/or medical procedures administered by an \emph{obstetrician} (or \emph{gynecologist}) are in general different from those by an \emph{oncologist}.
Thus we train the embedding of a doctor in an auxiliary task by predicting the doctor's primary specialty from his or her conducted medical services.
We initialize the embedding of a doctor as the weighted average of the embedding vectors of the medical services conducted by the doctor.

We use the Graph Attention Network \cite{velivckovic2017graph} to predict doctor specialties from services.
For a doctor $d_j$ whose conducted medical services are $\{s_i\}^{(d_j)}$, the normalized attention coefficient $\alpha_{ij}$ between the doctor embedding $\mathbf{d}_j$ and each of the service embeddings $\{\mathbf{s}_i\}^{(d_j)}$ conducted by doctor $d_j$ is
\begin{equation}
\alpha_{ij} = \frac{\text{exp}\left(\texttt{LeakyReLU}(\mathbf{a}^T[\mathbf{Wd}_j||\mathbf{Ws}_i])\right)}{\sum_{s_k \in \mathcal{N}_{d_j}}\text{exp}\left(\texttt{LeakyReLU}(\mathbf{a}^T[\mathbf{Wd}_j||\mathbf{Ws}_k])\right)}.
\label{eq:alpha}
\end{equation}
where $\{\mathbf{d, s}\} \in \mathbb{R}^p$, $\mathbf{a} \in \mathbb{R}^{2p^{\prime}}$, $\mathbf{W} \in \mathbb{R}^{p^{\prime} \times p}$, \texttt{LeakyReLU} is the Leaky Rectified Linear Unit with a negative input slope of 0.2 \cite{maas2013rectifier}, $\cdot^T$ represents transposition, and $\parallel$ is the concatenation operation.
$\{\mathbf{W, a}\}$ are parameters of the aggregation functions that ``aggregate'' the information of neighboring service vertices into the targeted doctor vertex.

The updated embedding vector of doctor $d_j$ can then be obtained as a linear combination of the associated service embeddings weighted by corresponding attention coefficients.
We adopt a $K$-head attention, such that the output dimension of the attention layer is $Kp^{\prime}$ instead of $p^{\prime}$:
\begin{equation}
\mathbf{d}_j^{\prime} = \parop_{k=1}^K \sigma \left(\sum_{s_i \in \mathcal{N}_{d_j}}\alpha_{ij}^k\mathbf{W}^k\mathbf{s}_i\right).
\label{eq:multi_head_att}
\end{equation}
Note that we have already obtained $\mathbf{s}_i$, thus making the doctor embedding a simper task than ordinary graph embedding wherein the embeddings of all nodes are unknown and to be learned.

\noindent\textbf{Patient Embedding}$\quad$
The similarity between patients can be defined from the perspectives of shared doctors and/or services.
In general, we expect the patient embedding can facilitate that \emph{patients are more similar to each other if they receive the same medical services from the same doctors.}

The versatile forms of patient similarity can be formalized as a \emph{bipartite multigraph} $\mathcal{G}_{pat}$, where the two disjoint sets of vertices ($\mathcal{P}$ and $\mathcal{S}$) represent the patients and services, respectively. 
A multigraph allows multiple edges connecting a node pair, which precisely models the scenario that a patient may have received the same service multiple times from different doctors.
An edge connecting patient $p_k$ and service $s_i$ carries two attributes: the doctor $d_j$ who treated $p_k$ with $s_i$, and the weight $w_{p_k \rightarrow d_j \rightarrow s_i}$ denoting the count of the service.

We propose a simple and scalable node embedding algorithm tailored for attributed multigraph based on \texttt{LINE} \cite{tang2015line}. 
We design a procedure \emph{duplication \& annotation} to convert $\mathcal{G}_{pat}$ into a simple graph with no attributes.
We first \emph{duplicate} each service node by the number of unique attributes of the edges linked to the node. 
A service node will not be duplicated if all its edges are of the same attribute. 
After \emph{duplication}, a service node must connect to either multiple edges with the same attribute or a single attributed edge.
We then \emph{annotate} each service node with the attribute of its edges, and remove the doctor attribute from its edges.
\emph{Annotation} can be implemented as a linear transformation of the concatenation of the doctor and service embedding vectors, which we have already obtained:
\begin{equation}
\mathbf{h}_{s_i,d_j} = \mathbf{W}_{a}[\mathbf{s}_i || \mathbf{d}_j]+\mathbf{b}_{a},
\end{equation}
where $\mathbf{W}_{a} \in \mathbb{R}^{p^{\prime\prime} \times (p+p^\prime)}$, $\mathbf{b}_{a} \in \mathbb{R}^{p^{\prime\prime}}$, and $\mathbf{h}_{s_i,d_j} \in \mathbb{R}^{p^{\prime\prime}}$ is the embedding of the new hybrid node.
Note that \emph{duplication \& annotation} will not significantly increase the computational overheads, as
(i) in a database of medical records, normally the number of patients is far greater than the numbers of doctors and medical services, and
(ii) a patient would frequently visit the same doctor for the same one or several services, hence the number of unique pairs of doctor-service is much smaller than the product of the numbers of doctors and medical services.

In \texttt{LINE}, node embeddings are optimized by preserving nodes' first-order and second-order proximities.
As in patient embedding, we are dealing with a bipartite graph, and that the embedding vectors of the hybrid nodes are already known (except for the transformation parameters), we can skip the first-order part and optimize the second-order part only.
For a patient $p_k$, its second-order proximity relative to other patients is defined over the ``context'' probability of seeing a hybrid node $h_{s_i, d_j}$:
\begin{equation}
p_2(h_{s_i, d_j}|p_k)=\frac{\text{exp}(\mathbf{h}_{s_i,d_j} \cdot \mathbf{p}_k)}{\sum_{l \in \{h\}}\text{exp}(\mathbf{h}_l \cdot \mathbf{p}_k)},
\label{eq:p2c}
\end{equation}
where $\mathbf{p}_k \in \mathbb{R}^{p^{\prime\prime}}$ and $\{h\}$ is the collection of all hybrid nodes.
Meanwhile, each context probability $p_2$ corresponds to an empirical distribution defined by the edge weights:
\begin{equation}
\hat{p}_2(h_{s_i, d_j}|p_k)=\frac{w_{p_k \rightarrow h_{s_i,d_j}}}{\sum_{l \in \mathcal{N}_{p_k}}w_{p_k \rightarrow h_l}},
\label{eq:p2e}
\end{equation}
where $\mathcal{N}_{p_k}$ represents the collection of all hybrid node neighbors of patient $p_k$.
Then we can optimize $\{\mathbf{p}_k\}_{k=1}^P$, $\mathbf{W}_{a}$, and $\mathbf{b}_{a}$ by minimizing the \emph{Kullback–Leibler} (\emph{KL}) distance between $\hat{p}_2$ and $p_2$:
\begin{equation}
\mathcal{L}_{pat} = -\sum_{(i, j, k) \in \mathcal{E}_{pat}} \frac{w_{p_k \rightarrow h_{s_i, d_j}}}{\sum_{l \in \mathcal{N}_{p_k}}w_{p_k \rightarrow h_l}}\text{log}(p_2(h_{s_i,d_j}|p_k)),
\end{equation}
where $\mathcal{E}_{pat}$ is the set of all edges of the patient-service bipartite graph after \emph{duplication \& annotation}.

\section{Experiments}
\vspace{-5pt}

\noindent\textbf{Experimental Setup}$\quad$
We test the proposed method on a proprietary clinical dataset that consists of medical records for patients who are either diagnosed as chronic lymphocytic leukemia (CLL) or undiagnosed as CLL but with related risk factors and/or symptoms.
The CLL-related risk factors and symptoms are pre-specified by a medical expert.
For CLL patients, we pulled their one-year medical records backward from six months before the date of diagnosis.

We compare \texttt{ME2Vec} with the following baselines for medical entity embedding:
\texttt{node2vec} \cite{grover2016node2vec}, \texttt{LINE}, spectral clustering (SC) \cite{ng2002spectral}, and non-negative matrix factorization (NMF) \cite{lee2001algorithms}.
For \texttt{ME2Vec}, the context window length $T$ is set as 8 days, and the number of attention heads $K$ is 4.
In practice, we found that the quality of learned medical entity embeddings is not sensitive to $T$, which can be chosen arbitrarily from 5 to 10.
The number of negative samples when training all methods is set as 10.
The dimensions of embeddings for all entities are set as 128.
The remaining parameter settings for all baselines are as default.

\begin{figure}[t!]
    \centering
    \begin{subfigure}[t]{0.5\textwidth}
        \centering
        \includegraphics[height=2in]{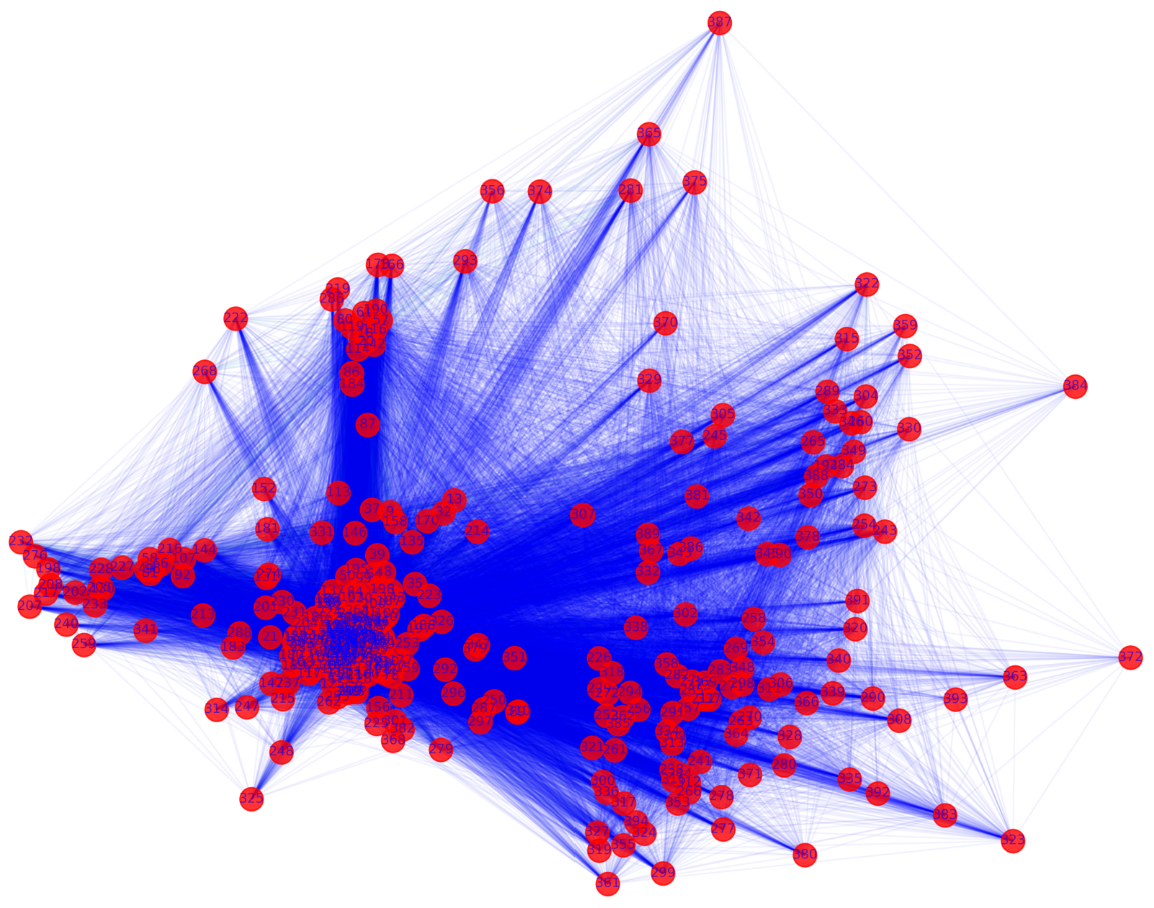}
    \end{subfigure}%
    ~ 
    \begin{subfigure}[t]{0.5\textwidth}
        \centering
        \includegraphics[height=2in]{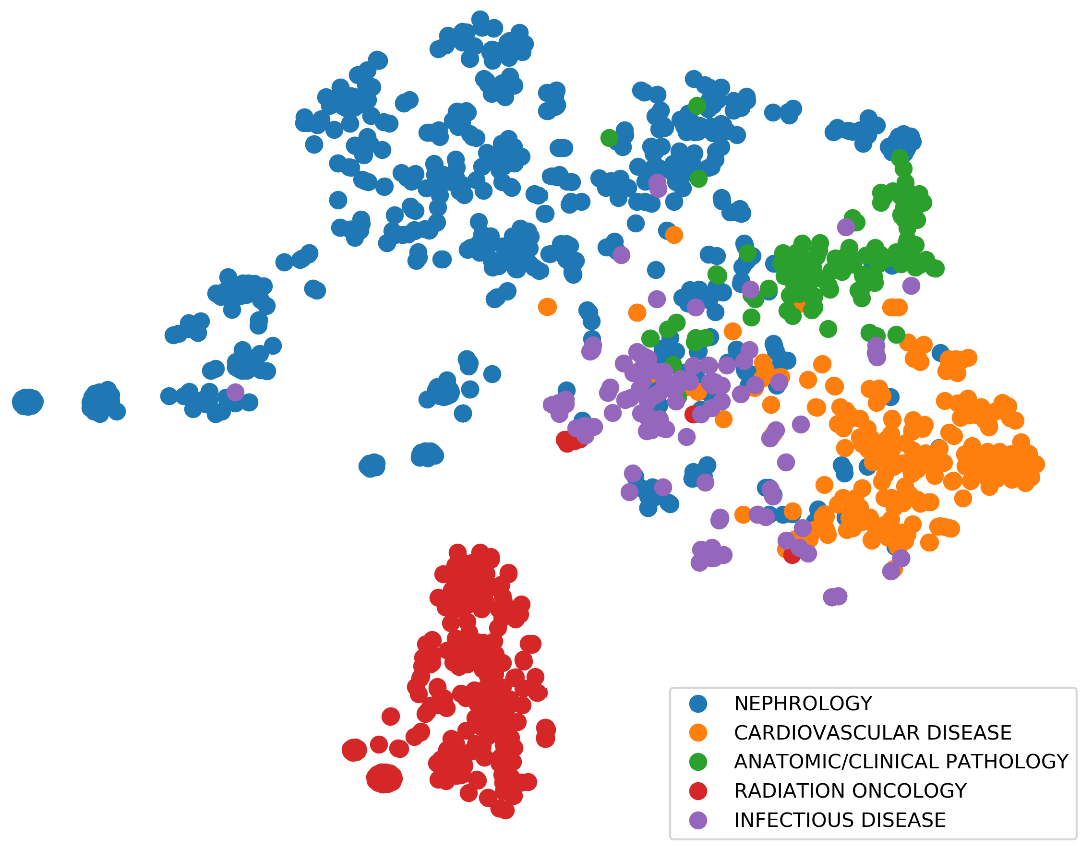}
    \end{subfigure}
    \caption{\small 2-dimensional visualization of service and doctor embeddings after PCA and t-SNE, respectively. Left: Each red dot represents a medical service with its ID labeled. Each blue line connecting two dots indicates that the two services co-occur as least once. Right: Each dot represents a doctor, with its color indicating the doctor's primary specialty. Doctors with five different primary specialties are displayed for illustration.}
    \label{fig:vis-emb}
    \vspace{-10pt}
\end{figure}

\noindent\textbf{Visualization of Service and Doctor Embedding}$\quad$
We visualize the trained embedding vectors of all medical services and some doctors in Figure \ref{fig:vis-emb}.
On the left part of Figure \ref{fig:vis-emb}, infrequent services (with larger IDs) spread out in the embedding space, whereas routine services (with smaller IDs) aggregate themselves closely in the centering area, which ensures the ``spatial isolation'' of important medical services.
On the right, we can see a clear separation of doctors with different primary specialties.
For example, \emph{nephrology} doctors are far away from \emph{cardiovascular disease} doctors, while \emph{radiation oncology} doctors are even further away from the rest.

\noindent\textbf{Node Classification}$\quad$
We first train \texttt{ME2Vec} and the baselines on the entire dataset to obtain patient embeddings for each of the methods.
Unlike \texttt{ME2Vec}, the baselines cannot integrate information from both doctors and services at the same time.
To address this, we create two bipartite graphs from the dataset that model the patient-doctor and patient-service relations, respectively.
Therefore each baseline has two versions of patient embeddings, with one learned from the patient-service graph, and the other learned from the patient-doctor graph.
We tried simply concatenating the two versions of embeddings, however the performance was no better than using them separately, thus not reported.

Next, we use the patient embeddings in the training set as well as their CLL diagnostic labels to train a logistic regression (LR) classifier with L2 regularization.
After that, we predict the diagnostic labels of patients in the testing set from their embeddings using the trained LR classifier.
We vary the training ratio from 20\% to 80\%, and under each training ratio we repeat the experiment for 10 times with randomized train/test split and report the average Micro-F1 and Macro-F1 in Table \ref{tb:microf1}.
The results show that \texttt{ME2Vec} outperforms all the baselines.
All the baselines achieve consistently poorer performance on the patient-doctor graph, suggesting their common weakness of extracting useful information from the patient-doctor relation.

\begin{table}[t]
\footnotesize
\caption{Performance of node classification in micro-F1 and macro-F1.}
\centering
\begin{tabular}{c c c c c c c c c}
\toprule[.8pt]
\multirow{2}{*}{Algorithms} & \multicolumn{4}{c}{Micro-F1} & \multicolumn{4}{c}{Macro-F1} \\
\cmidrule(r){2-5}
\cmidrule(r){6-9}
& 20\% & 40\% & 60\% & 80\% & 20\% & 40\% & 60\% & 80\% \\
\midrule[.3pt]
ME2Vec & \textbf{0.869} & \textbf{0.877} & \textbf{0.878} & \textbf{0.879} & \textbf{0.664} & \textbf{0.679} & \textbf{0.682} & \textbf{0.676} \\
\midrule[.3pt]
node2vec (service) & 0.865 & 0.875 & 0.876 & 0.878 & 0.613 & 0.630 & 0.632 & 0.640 \\
node2vec (doctor) & 0.850 & 0.862 & 0.860 & 0.861 & 0.474 & 0.466 & 0.462 & 0.463 \\
\midrule[.3pt]
LINE (service) & 0.855 & 0.864 & 0.866 & 0.866 & 0.587 & 0.592 & 0.592 & 0.586 \\
LINE (doctor) & 0.854 & 0.863 & 0.860 & 0.861 & 0.470 & 0.465 & 0.462 & 0.463 \\
\midrule[.3pt]
SC (service) & 0.862 & 0.861 & 0.861 & 0.868 & 0.463 & 0.463 & 0.463 & 0.465 \\
SC (doctor) & 0.862 & 0.861 & 0.861 & 0.868 & 0.463 & 0.463 & 0.463 & 0.465 \\
\midrule[.3pt]
NMF (service) & 0.868 & 0.870 & 0.869 & \textbf{0.879} & 0.584 & 0.586 & 0.589 & 0.600 \\
NMF (doctor) & 0.861 & 0.860 & 0.860 & 0.867 & 0.469 & 0.472 & 0.470 & 0.469 \\
\bottomrule[.8pt]
\label{tb:microf1}
\end{tabular}
\vspace{-20pt}
\end{table}

\section{Conclusions}
\vspace{-5pt}

In this paper, we propose a unified and hierarchical medical entity embedding framework \texttt{ME2Vec} for representation learning of EHR data.
We design a time-aware service embedding that can leverage the temporal profiles of medical services to characterize their importance towards evaluating patient similarity.
Moreover, we develop an effective approach of node embedding for attributed multigraph that uniquely addressed the difficulty of patient embedding learning from both doctors and services.
We conduct experiments on a real-world clinical dataset, and show that \texttt{ME2Vec} outperforms strong baselines, thanks to its unified and hierarchical structure of information fusion.

\end{document}